\documentclass{article}
\usepackage{ijcai23}

\usepackage{times}
\usepackage{soul}
\usepackage{url}
\usepackage[hidelinks]{hyperref}
\usepackage[utf8]{inputenc}
\usepackage[small]{caption}
\usepackage{graphicx}
\usepackage{amsmath}
\usepackage{amsthm}
\usepackage{booktabs}
\usepackage{algorithm}
\usepackage{algorithmic}
\usepackage[switch]{lineno}

\title{Elucidating STEM Concepts through Generative AI: A Multi-modal Exploration of Analogical Reasoning}

\author{
Chen Cao$^{1,2}$
\and
Zijian Ding$^{3,4}$\and
Gyeong-Geon Lee$^5$\and
Jiajun Jiao$^6$\and
Jionghao Lin$^7$\and
Xiaoming Zhai$^5$
\affiliations
$^1$Carnegie Learning\\
$^2$University of Sheffield\\
$^3$Microsoft Research\\
$^4$University of Maryland,
College Park\\
$^5$University of Georgia\\
$^6$New York University\\
$^7$Carnegie Mellon University
\\
\vspace{2mm}
\emails
ccao5@sheffield.ac.uk,
ding@umd.edu,
crusaderlee@snu.ac.kr,
jj3100@nyu.edu,
jionghao@cmu.edu,
xiaoming.zhai@uga.edu
}

\begin{document}

\maketitle

\begin{abstract}
This study explores the integration of generative artificial intelligence (AI), specifically large language models, with multi-modal analogical reasoning as an innovative approach to enhance science, technology, engineering, and mathematics (STEM) education. We have developed a novel system that utilizes the capacities of generative AI to transform intricate principles in mathematics, physics, and programming into comprehensible metaphors. To further augment the educational experience, these metaphors are subsequently converted into visual form. Our study aims to enhance the learners' understanding of STEM concepts and their learning engagement by using the visual metaphors. We examine the efficacy of our system via a randomized A/B/C test, assessing learning gains and motivation shifts among the learners. Our study demonstrates the potential of applying large language models to educational practice on STEM subjects. The results will shed light on the design of educational system in terms of harnessing AI's potential to empower educational stakeholders.\end{abstract}

\section{Introduction}

The teaching and learning of science, technology, engineering, and mathematics (STEM) stand as an essential yet challenging aspect of education. Concepts, the bedrock of STEM, are frequently abstract and complex, demanding learners to exhibit higher-order thinking skills such as logical reasoning and problem-solving. Consequently, comprehending these algorithms often poses significant hurdles for learners, particularly for those new to the field \cite{lee2011personifying}.

Current pedagogical practices primarily rely on textual explanations and code examples to convey these concepts, which, although useful, can be limiting. Traditional methods might fail to demystify abstract concepts adequately, relying heavily on learners' abilities to translate theoretical knowledge into functional understanding \cite{kolikant2010digital}. Moreover, the inherently linear nature of text-based teaching methods may not cater well to the non-linear and dynamic nature of STEM concepts \cite{sorva2013review}.

Educators, particularly in the STEM fields, face their own set of challenges. Lesson planning, particularly for complex topics like mathematical theorems, physical laws, and computational algorithms, can be a daunting task, demanding not just extensive subject knowledge, but also creativity in crafting engaging content \cite{Xu2022TheAO}. Resource constraints, such as limited time and funding, further add to these challenges. Technological limitations also come into play, as the tools to create interactive, multimodal learning content are often expensive or difficult to use \cite{lee2014principles}.

Recognizing these challenges, this paper introduces an innovative, AI-driven, multimodal approach to teaching STEM algorithms. The proposed approach harnesses the capabilities of a state-of-the-art large language model, to generate intuitive analogies and transform them into engaging visual storyboards with a text-to-image generative model. The aim is to enhance pedagogical efficiency and, most importantly, to improve learners' comprehension of complex STEM algorithms.


\section{Related Work}





\subsection{AI for Analogical Reasoning}

Analogical reasoning, a cognitive process that involves transferring knowledge from a source domain to a target domain based on similarities and differences, has long been recognized as a powerful tool for problem-solving and learning \cite{gentner1997structure,holyoak1996mental}. Analogies can help learners make connections between familiar and unfamiliar concepts, facilitating understanding and promoting knowledge transfer \cite{richland2007cognitive}. Recent advancements in AI, particularly the development of large language models like GPT-4, have unlocked new possibilities for generating analogies and facilitating their integration into educational contexts. Previous research suggests that Large Language Models (LLMs) might have the ability to create analogies that are similar to those made by humans, including longer natural-language analogies \cite{teamWorldCreationAnalogy2020}, explain analogical mappings \cite{webbEmergentAnalogicalReasoning2022,bhavyaAnalogyGenerationPrompting2022}, or invent analogy-inspired concepts for creative problem-solving \cite{zhuGenerativePreTrainedTransformer2022,zhuBiologicallyInspiredDesign2023,webbEmergentAnalogicalReasoning2022,ding2023fluid,Ding2023MappingTD}.

\subsection{Analogical Reasoning to Support Education}

Previous work highlighted how AI can generate personalized learning pathways for students, leading to enhanced learning outcomes \cite{tetzlaff2021developing}, and underscored the potential of AI in predicting and improving student engagement and performance \cite{maghsudi2021personalized}.


One strength of AI is the ability to simplify complex concepts. For STEM topics, it can generate relevant analogies or stories that draw on more familiar situations or narratives. For example, variables in STEM can be thought of as boxes where you store things (values), which you label so you know what's inside \cite{tetzlaff2021developing}. Additionally, for visual learners, storyboards or visual analogies can be created to help explain these concepts. Adaptive storytelling refers to tailoring the narrative based on the individual's understanding or preference for learning, which can be gauged through their interactions or explicit feedback \cite{bietti2019storytelling}.
This ability can be employed to generate study materials, simplifying educators' workload and enhancing the learning experience.

\subsection{STEM Education}
Learning to program is increasingly recognized as a critical skill. However, acquiring this skill can be a daunting task due to the complexity and abstractness of STEM algorithms \cite{selby2015relationships,marin2018empirical}. Despite numerous pedagogical strategies—like pair STEM, code tracing, and the use of visual aids—many students still find it challenging to comprehend and apply algorithmic concepts \cite{kalelioglu2014effects}.

A common pedagogical strategy for teaching algorithms is through textual explanations and examples \cite{lopez2008relationships}. However, this approach can be insufficient, particularly for novices, as it requires the learner to translate the abstract, textual information into a more concrete understanding \cite{kolikant2010digital}. This translation process is often fraught with challenges and can result in misconceptions.

\subsection{Multimodal Learning and Cognitive Theory of Multimedia}

Given the challenges of textual explanations, researchers have begun exploring multimodal learning approaches, which combine text with other modalities, such as images or videos. Mayer's \cite{mayer2002multimedia} cognitive theory of multimedia learning posits that people learn better from words and pictures together than from words alone. His research has sparked a growing interest in multimodal learning in various fields, including computer science education \cite{betrancourt2005animation}.

Aligned with Mayer's theory, Paivio's \cite{paivio1991dual} dual coding theory proposes that verbal and non-verbal systems are used for cognitive processing. He suggests that information can be remembered better if it is presented in both verbal and visual formats. This theory has been used to inform many multimodal learning approaches, which attempt to exploit both the visual and verbal cognitive subsystems.

\subsection{Synthesis and Research Gap}

The reviewed literature shows the potential of AI in education and the need for innovative approaches to teaching STEM algorithms. There is also theoretical support for using multi-modal approaches to enhance learning. However, there is a clear gap in research that combines these elements—AI, multi-modal learning, and STEM education. This study aims to fill this gap by exploring an AI-driven, multi-modal approach to teaching STEM algorithms.

In this paper, we present an innovative approach that harnesses the power of analogical reasoning, multimodal learning, and AI technologies to demystify STEM concepts. By combining AI-generated analogies with visual storyboards and animated videos, we aim to enhance learners' comprehension and knowledge retention in the field of STEM. We believe that this multimodal approach has the potential to revolutionize STEM education, inspire creative instructional content creation, and shed light on the interplay between analogical reasoning, abstraction, and cognitive processes.

This innovative educational approach sits at the intersection of AI and education, an increasingly researched area that holds immense potential. Leveraging AI to aid teaching and learning can provide personalized learning experiences, making education more effective and engaging. Therefore, an exploration into the role of AI, especially in generating analogies and visual aids, contributes valuable insights into this burgeoning field.

In addition to its theoretical implications, this research has practical significance. If successful, our approach could be a valuable tool for educators, aiding them in planning engaging lessons and providing students with a unique, effective way to understand abstract STEM concepts. In the long run, it could significantly transform teaching and learning processes in computer science and beyond.

The potential of AI in education has been well-documented, with significant strides in personalized learning and content generation. However, the application of AI for generating intuitive analogies and visual aids in STEM education is an under-explored domain. Our approach aims to fill this gap, exploiting the power of AI to create an engaging learning experience.

\section{Theoretical Framework}

This study is underpinned by three major theoretical perspectives: Mayer's cognitive theory of multimedia learning, Paivio's dual coding theory, and Piaget's theory of cognitive development.

Mayer \cite{mayer2002multimedia} and Paivio \cite{paivio1991dual} provide the rationale for employing a multimodal approach, suggesting that learners process information more effectively when presented using both verbal and visual means. Piaget's theory of cognitive development \cite{piaget1970science} further supports our approach. Piaget postulates that knowledge acquisition is an active, constructive process where learners constantly build and adjust their understanding by integrating new information with their existing knowledge base. This theory justifies our approach's interactive nature, which prompts learners to actively engage with the content, facilitating the integration of new information.

Our methodology is also heavily informed by best practices in UX/UI design. Following Norman's principles \cite{norman1988psychology}, the interface is designed to be simple and intuitive, minimizing the cognitive load on the users. Clear instructions and feedback mechanisms are incorporated to guide the users through the process of selecting an analogy, generating a storyboard, and creating the animated video. The design also allows flexibility for the users to edit the generated content, providing a sense of control and customization, which further enhances the user experience.

These theories collectively inform the design of our AI-driven, multimodal learning solution that leverages analogies and visual storyboards. We hypothesize that this approach can enhance the comprehension of complex STEM algorithms by stimulating multiple cognitive subsystems and fostering an active learning environment.

\section{Methodology}

We developed a prototype with live deployment\footnote{https://analogen.netlify.app/}. The prototype is backed by GPT-4, one of the most advanced language models developed by OpenAI (in Jul 2023), to generate a set of analogies based on specific STEM concepts. Users can provide a STEM concept (e.g., ``Newton's first Law'') as input, and the GPT-4 model generates three distinct analogies (e.g., skating on ice, pushing a stalled car, and the stationary soccer ball) as shown in the Appendix \ref{appendix}. These analogies are designed to be intuitive and engaging, catering to a range of learner preferences.

Following the analogy selection, the system uses preset prompts to create a narrative based on the chosen analogy. This narrative serves as the foundation for generating a storyboard. The storyboard comprises a sequence of four images and each iamge is accompanied by a description, which provides a step-by-step visualization of the STEM concept using the selected analogy.

Finally, the images and descriptions from the storyboard are transformed into an animated video. The video serves as an engaging and easily digestible medium to convey the STEM concept building upon the principles of multimedia learning and dual coding theories.

\section{Preliminary Results}

We examined the content generation performance of our tool through a sequence of incremental steps, each substantiated by specific cases. These steps include:
\begin{itemize}
  \item Transforming STEM concepts into text-based analogies.
  \item Converting text-based analogies into static visual analogies.
  \item Evolving static visual analogies into dynamic visual analogies.
\end{itemize}
In the initial phase, our trials revealed that our tool was proficient in generating informative text-based analogies from STEM concepts. For instance, in the context of Object-Oriented Programming (OOP), objects were analogized as Lego bricks, and classes were likened to structures of Lego.

In the second phase, our tool was capable of producing visually appealing images for text-based analogies. However, we encountered a challenge in that STEM concepts often encompass multiple components that must be depicted in an analogy. This complexity is evident in analogies for OOP (objects, classes) and the analogy between water pressure/current and electric voltage/current in physics. Despite our emphasis on differentiating these components in the image generation prompts, the generative model frequently fell short in representing each part comprehensively. For example, consider the prompt for generating the analogy between water pressure/current and electric voltage/current:
\begin{quote}
  "Water flows through a narrow tube connected between two water tanks, one tank having significantly more water than the other."
\end{quote}

\begin{figure}[htp]
    \centering
    \includegraphics[width=0.46\textwidth]{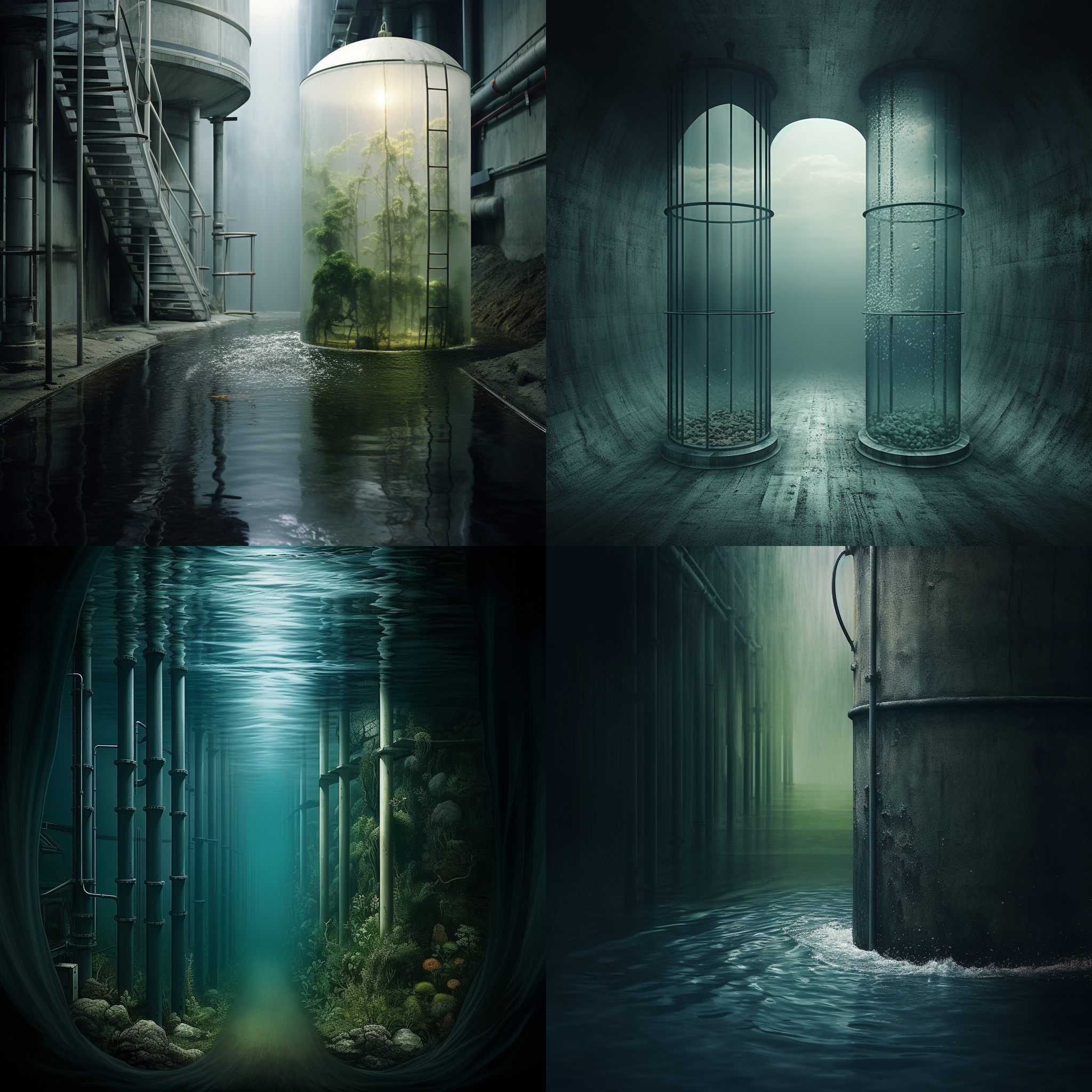}
    \caption{Visual analogy for water pressure/current and electric voltage/current with prompt "Water flows through a narrow tube connected between two water tanks, one tank having significantly more water than the other." Only the image in the top right corner (the second one) displayed two water tanks, with one tank (right) containing more water than the other. However, it was missing a connecting tube, an important feature needed to symbolize the relationship between water current and electric current.}
    \label{fig:example}
\end{figure}

From the generated images as shown in Figure \ref{fig:example}, only the second one (in the top right corner) featured two water tanks, with one having more water than the other, but it lacked a connecting tube, a critical element to represent the mapping between water current and electric current.

In the third phase, the creation of dynamic visual analogies, while visually appealing, posed challenges in devising suitable transitions or motions to articulate the analogy dynamically. Taking the analogy between water pressure/current and electric voltage/current as an example, we found it difficult to generate a visual representation that effectively conveyed the concept of "water flowing from the tank with more water to the tank with less water."



\section{Discussion}

While the preliminary results indicate promising outcomes in terms of content generation and potential educational effectiveness, it is essential to recognize the need of an empirical study with learner participants in a real-world setting. 

 
Our next step is to conduct a between-subject study with different types of analogies to gain valuable insights into the actual learning outcomes, learner engagement, and potential areas for improvement. Additionally, expanding the study to different STEM concepts and considering diverse learner profiles could help assess the generalizability and effectiveness of the approach in various educational settings.

\section{Conclusion}
This paper introduces a generative-AI-facilitated pedagogical tool for computer science education. Our results reveal the potential of expanding this multimodal approach to other subjects (e.g., physics, chemistry, and mathematics) in future research. By revolutionizing the creation of educational content, we aim to break down the barriers to understanding complex leanring concepts and foster an environment conducive to both teaching and learning.

\bibliographystyle{ACM-Reference-Format}
\bibliography{ijcai23}

\onecolumn

\appendix
\textbf{APPENDIX}

\section{Workflow and example interface for multi-media analogy generation}
\label{appendix}

\begin{figure*}[htp]
    \centering
    \includegraphics[width=0.6\textwidth]{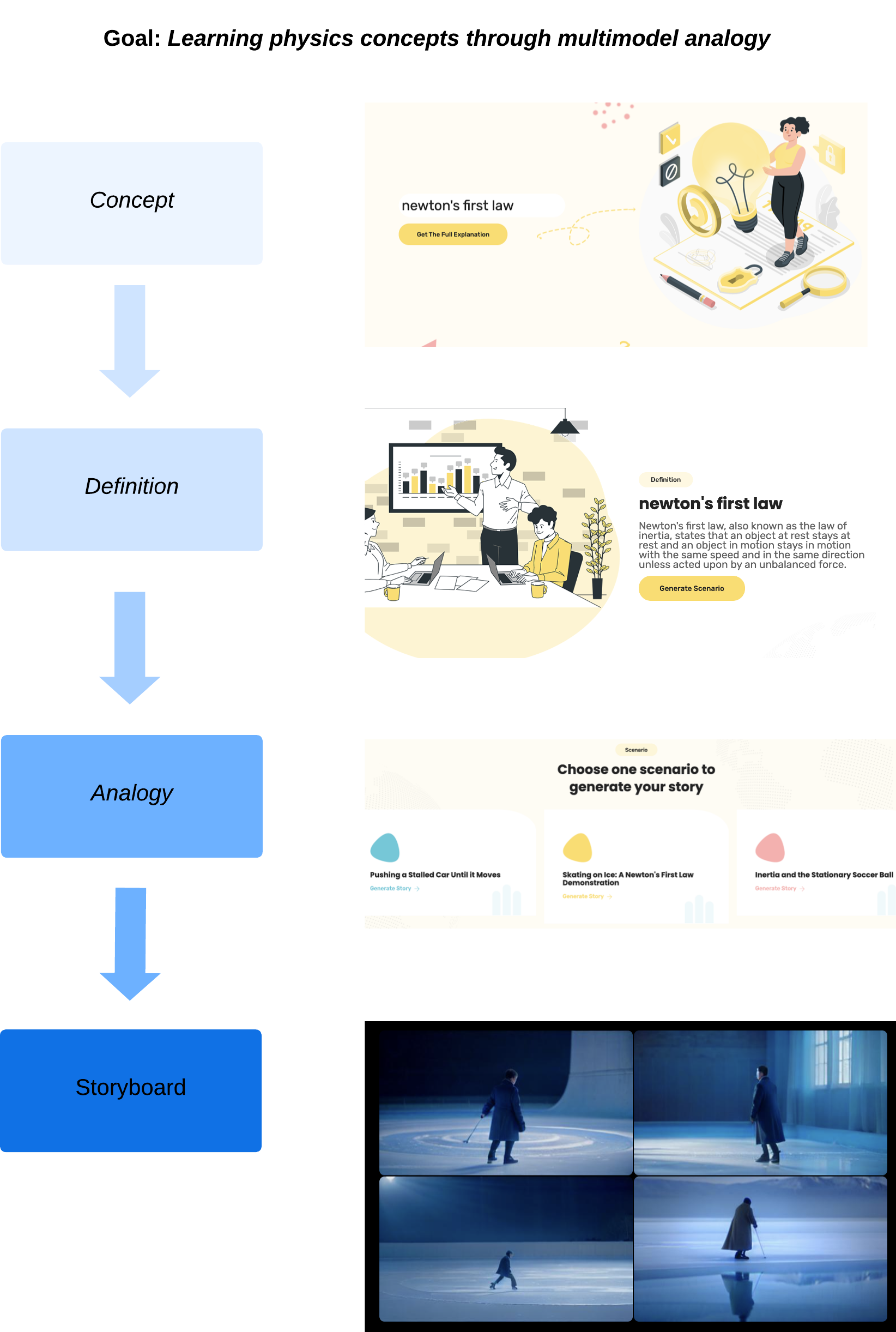}
    \caption{Initiate a process by inputting a specific STEM concept like ``Newton's First Law''. Validate its definition, followed by generating three distinct analogies for comprehensive understanding. Subsequently, select a preferred analogy to create a detailed, four-scene storyboard, complete with vivid images and thorough descriptions. Lastly, transform these dynamic scenes into an engaging instructional video.}
    \label{fig:flowchat}
\end{figure*}

\end{document}